\pdfoutput=1

\documentclass[11pt]{article}

\usepackage[]{acl}

\usepackage{multirow}
\usepackage{graphicx}
\usepackage{booktabs}
\usepackage{tabularx}
\usepackage{CJKutf8}
\usepackage{amssymb}
\usepackage{amsmath}
\usepackage{tabu} 
\usepackage{hyperref}
\usepackage{breakurl}
\usepackage{times}
\usepackage{latexsym}
\usepackage{tikz}
\usepackage{pgfplots}
\usepackage[normalem]{ulem}
\usepackage{xcolor}
\usepackage{dashrule}
\usepackage{color, colortbl}

\usepackage{pifont}
\newcommand{\cmark}{\ding{51}}
\newcommand{\xmark}{\ding{55}}
\definecolor{brickred}{HTML}{b92622}
\definecolor{midnightblue}{HTML}{005c7f}
\definecolor{salmon}{HTML}{f1958d}
\definecolor{burntorange}{HTML}{f19249}
\definecolor{junglegreen}{HTML}{4dae9d}
\definecolor{forestgreen}{HTML}{499c5e}
\definecolor{pinegreen}{HTML}{3d8a75}
\definecolor{seagreen}{HTML}{6bc1a2}
\definecolor{limegreen}{HTML}{97c65a}

\usepackage{lipsum}

\newcommand\blfootnote[1]{%
  \begingroup
  \renewcommand\thefootnote{}\footnote{#1}%
  \addtocounter{footnote}{-1}%
  \endgroup
}

\usepackage[T1]{fontenc}

\usepackage[utf8]{inputenc}

\usepackage{microtype}

\title{Mining Error Templates for Grammatical Error Correction}

\author{Yue Zhang$^{1\dagger}$, Haochen Jiang$^1$, Zuyi Bao$^2$, {\bf Bo Zhang$^2$},   {\bf Chen Li$^2$},  {\bf Zhenghua Li$^{1}$} \\
        $^1$Institute of Artificial Intelligence, School of Computer Science and Technology, \\
Soochow University, China; $^2$DAMO Academy, Alibaba Group, China\\
\texttt{$^1$\{yzhang21,hcjiang\}@stu.suda.edu.cn}, \texttt{$^1$zhli13@suda.edu.cn}\\\texttt{$^2$\{zuyi.bzy,klayzhang.zb,puji.lc\}@alibaba-inc.com}}

\begin{document}
\begin{CJK}{UTF8}{gkai}
\maketitle

\begin{abstract}
Some grammatical error correction (GEC) systems incorporate hand-crafted rules and achieve positive results.
However, manually defining rules is time-consuming and laborious.
In view of this, we propose a method to mine error templates for GEC automatically. An error template is a regular expression aiming at identifying text errors. We use the web crawler to acquire such error templates from the Internet. For each template, we further select the corresponding corrective action by using the language model perplexity as a criterion. We have accumulated 1,119 error templates for Chinese GEC based on this method. Experimental results on the newly proposed CTC-2021 Chinese GEC benchmark show that combing our error templates can effectively improve the performance of a strong GEC system, especially on two error types with very little training data.\footnote{Our error templates are available at \url{https://github.com/HillZhang1999/gec_error_template}.}
\end{abstract}
\section{Introduction}
Grammatical error correction (GEC) is an important task in natural language processing, which aims at detecting and correcting all underlying errors in a potentially erroneous sentence. Recently, GEC has been receiving increasing attention for its broad application \citep{grundkiewicz2020crash,wang2021comprehensive}. \blfootnote{$^\dagger$ This work was partially done during the first author's internship at Alibaba DAMO Academy.}

Early GEC systems are basically based on error-specific classifiers \citep{rozovskaya2011algorithm,dahlmeier2012beam} or statistic machine translation models \citep{felice2014grammatical,chollampatt2016neural}. Since the beginning of the deep learning era, neural encoder-decoder models, e.g., Transformer \citep{vaswani2017attention}, have emerged as a dominant GEC paradigm \citep{yuan2016grammatical,junczys2018approaching}. Despite their discrepancies, all above methods need to train a model over numerous training samples through the back-propagation algorithm, so we can collectively call them model-based approaches. Besides model-based approaches, there is yet another simple but effective long-standing GEC approach, i.e., the rule-based method, which utilizes pre-defined rules to tackle grammatical errors \citep{madi2018grammatical}. Compared with the model-based approaches, the rule-based method enjoys some merits, such as 1) 
fast correction speed; 2) good interpretability and controllability; 3) no training data required, making this method widely applied in various languages \citep{domeij2000granska,sidorov2013syntactic,singh2016frequency,zhou2018chinese}. Moreover, some researchers attempt to combine the rule-based method with the model-based approaches and show that they have complementary abilities \citep{felice2014grammatical, zhang2021ctcreport}.

\begin{table}[tp!]
\centering
\scalebox{1.0}{
\begin{tabular}{cc}
\hline
\multicolumn{2}{c}{\textbf{Meaning duplicates}}   \\ \hline
\multirow{2}{*}{\textbf{Src.}} & 他大约五岁左右。                                                \\
                                    & He is about five years old or so.                                      \\ 
                                    \hline
\multirow{2}{*}{\textbf{Ref.}} & 他大约五岁\sout{左右}。                                                \\
                                    & He is about five years old \sout{or so}.                                      \\ \hline\hline
                                    \multicolumn{2}{c}{\textbf{Mixed sentence patterns}}   \\ \hline
\multirow{2}{*}{\textbf{Src.}}    & 杀人事件的起因是因为打牌争执。                                                   \\
                                    & \begin{tabular}[c]{@{}c@{}}The cause of the murder was \\ because a dispute over playing cards.\end{tabular} \\ \hline
\multirow{2}{*}{\textbf{Ref.}} & 杀人事件的起因是\sout{因为}打牌争执。                                                   \\
                                    & \begin{tabular}[c]{@{}c@{}}The cause of the murder was \\ \sout{because} a dispute over playing cards.\end{tabular} \\ \hline
\end{tabular}
}

\caption{Examples of two error types in CTC-2021 Chinese GEC dataset that can be well solved by our error templates. ``Src'' and ``Ref'' refer to source and reference, respectively. The \sout{strikethrough} means that delete words.}
\label{tab:example}
\end{table}
Despite its advantages, the rule-based GEC method is less commonly used today, mainly due to the additional labor costs. A grammar rule for GEC formally consists of two parts: 1) an error template to match erroneous spans; 2) a corrective action to correct errors. Existing work usually resorts to human experts to manually define such templates and actions, which inevitably introduces extensive expenses. These extra expenses consequently limit the scale of rules and make their effect marginal. 

\begin{figure}[tp!]
\centering
\includegraphics[scale=0.45]{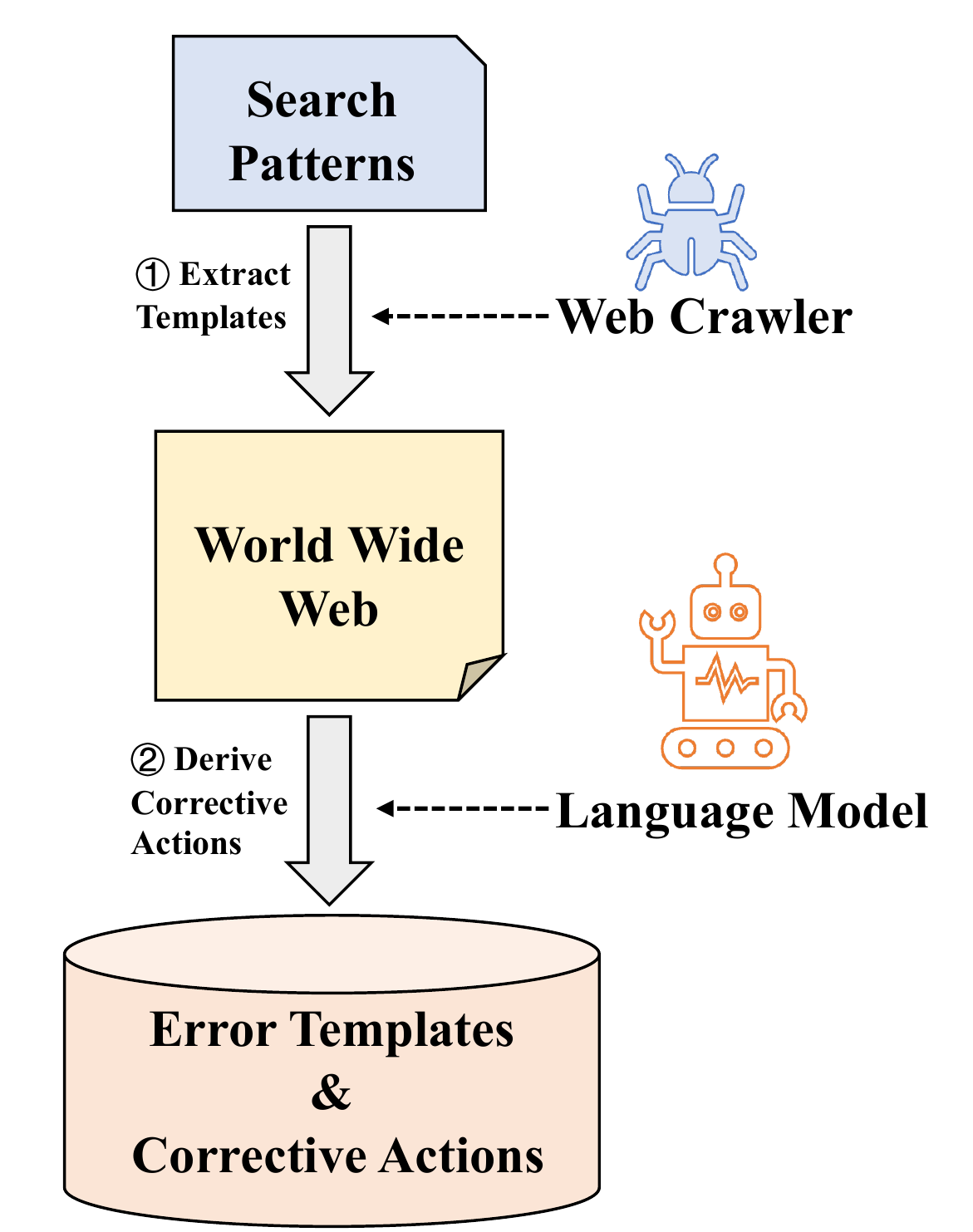}
\caption{The workflow of the proposed method. 
}
\label{fig:sp}
\end{figure}

To make the rule-based approach more feasible, we propose to automatically mine error templates from websites and devise a method to obtain corrective actions based on language models. The motivation is to minimize human involvement and thus reduce labor. In this work, we focus on the redundant error, which can be easily solved by directly deleting erroneous tokens. Concretely, we represent error templates as regular expressions in the form of ``A.*B'' and utilize a web crawler to extract such templates from the Internet based on several pre-defined search patterns. Then we design three corrective actions: deleting the left part ``A'', deleting the right part ``B'', and randomly deleting one part. To choose an appropriate action, we leverage the perplexity calculated by language models like GPT-2 \citep{radford2019language} as a criterion. 

We conduct experiments on the Chinese GEC task. So far, we have accumulated 1,119 Chinese error templates along with their corrective actions by using our proposed method. Experimental results on the newly proposed CTC-2021 benchmark\footnote{\url{https://github.com/destwang/CTC2021}} show that incorporating our templates leads to a significant improvement of 4.50 F${_1}$ score over a strong baseline, which confirms the effectiveness. Specifically, our error templates perform well on two special error types with 
very few training samples: \textit{meaning duplicates} and \textit{mixed sentence patterns}. As shown in Table \ref{tab:example}, both kinds of errors fit our templates well and can be fixed by the delete operation. Using our templates improves the recall values from 9.59 and 7.41 to 63.01 and 40.74 for \textit{meaning duplicates} and \textit{mixed sentence patterns}, respectively.

\section{Method}
\label{sec:met}
\begin{table}[tp!]
\centering
\scalebox{1.0}{
\begin{tabular}{cc}
\hline
\textbf{ID}                 & \textbf{Search Pattern}         \\ \hline
\multirow{2}{*}{\textbf{1}} & A...B是语法错误吗？                    \\
                            & Is A ... B a grammatical error? \\ \hline
\multirow{2}{*}{\textbf{2}} & A...B是病句吗？                      \\
                            & Is A ... B a ill sentence? \\ \hline
\multirow{2}{*}{\textbf{3}} & A...B是语义重复吗？                      \\
                            & Is A ... B a meaning redundant? \\ \hline
\multirow{2}{*}{\textbf{4}} & A...B是句式杂糅吗？                      \\
                            & Is A ... B a mixed sentence pattern? \\ \hline
\multirow{2}{*}{\textbf{5}} & A...B这句话错了吗？                      \\
                            & Is the sentence A ... B wrong? \\ \hline
\end{tabular}
}
\caption{Five search patterns used in our experiments.}
\label{tab:pattern}
\end{table}

In this section, we describe how to automatically obtain error templates and their corrective actions. We first use web crawlers to extract error templates from the Internet, then utilize language models to derive their corrective actions. The overview of our method is depicted in Figure \ref{fig:sp}. We have made a preliminary attempt in Chinese, and our method can be easily extended to other languages.

\subsection{Template Extraction}

To extract candidate error templates ``A.*B'' from the web, we design several search patterns.
Search patterns are sentence structures manually summarized from the common problems of people asking for grammatical errors, e.g., ``A...B是语法错误吗?'' (is A...B a grammatical error?). We use regular expressions to represent such search patterns and utilize them to match questions from the Q\&A platforms
by web crawlers. For example, we can derive an error template ``因为.*为由'' (since .* as a reason) from the question ``因为...为由是语法错误吗?'' (is ``since ... as a reason'' a grammatical error?). At present, we have used five search patterns, which can be expanded in the future. We show them in Table \ref{tab:pattern}.

We match about 2K questions using search patterns from  widely-used Chinese Q\&A platforms like \emph{Baidu Zhidao}\footnote{\url{{https://zhidao.baidu.com/}}}. Then, we automatically obtain candidate error templates from the reserved slots ``A'' and ``B'' in search patterns. We also manually adjust some templates as current search patterns can not always hit the target accurately. After removing duplicate templates, we get 1,119 Chinese error templates in total.

\paragraph{Human Evaluation.}
We launch a human evaluation to study the quality of extracted error templates. We randomly select 100 templates and ask two native Chinese speakers to annotate the acceptability of each template individually. The acceptability follows a 3-point scale:
\vspace{-3mm}
\begin{itemize} 
    \setlength{\itemsep}{-2pt}
    \setlength{\topsep}{-2pt}
    \item \textbf{[0]}: the template can \textit{hardly} match erroneous spans correctly;
    \item \textbf{[1]}: the template can \textit{sometimes} match erroneous spans correctly;
    \item \textbf{[2]}: the template can \textit{always} match erroneous spans correctly;
\end{itemize}
\vspace{-3mm}

\begin{table}[tp!]
\centering
\scalebox{1.0}{
\begin{tabular}{cc}
\hline
\textbf{Template}                                     & \textbf{Action}             \\ \hline
每天.*日理万机                                     & \multirow{2}{*}{ACT$_{left}$} \\
\begin{tabular}[c]{@{}c@{}}Everyday .* handle \\ numerous affairs everyday\end{tabular} &                    \\ \hline
原因是.*引起的                                     & \multirow{2}{*}{ACT$_{right}$} \\
\begin{tabular}[c]{@{}c@{}}The reason is .* cause\end{tabular} &                    \\ \hline
大约.*左右                                     & \multirow{2}{*}{ACT$_{random}$} \\
\begin{tabular}[c]{@{}c@{}}about .* or so\end{tabular} &                    \\ \hline
\end{tabular}
}
\caption{Examples of three corrective actions.}
\label{tab:act}
\end{table}
Figure \ref{fig:dis1} shows the human evaluation results. The average scores from annotators 1 and 2 are 1.38 and 1.47, respectively. The annotation consistency ratio is 63\%, and most disagreements arise in determining whether a template is 1 or 2 points since the boundary between \textit{always} and \textit{sometimes} is relatively vague. We then ask the two annotators to discuss and handle the inconsistent samples (Final). We can see that both annotators consider that 97\% of our templates can work in certain cases (score > 0) and 49\% of them can work in most cases (score = 2), which clearly presents the satisfactory quality of our templates.

\subsection{Corrective Action Acquisition}
The erroneous spans detected by our error templates can be corrected through three corrective actions: 1) ACT$_{left}$: deleting the left part of the template; 2) ACT$_{right}$: deleting the right part; 3) ACT$_{random}$: randomly selecting one side to delete. Table \ref{tab:act} shows examples for these actions. There are also some error templates that need other actions like substitution and insertion to correct. We leave such templates to our future work.

\begin{figure}[tp!]
\centering
\includegraphics[scale=0.55]{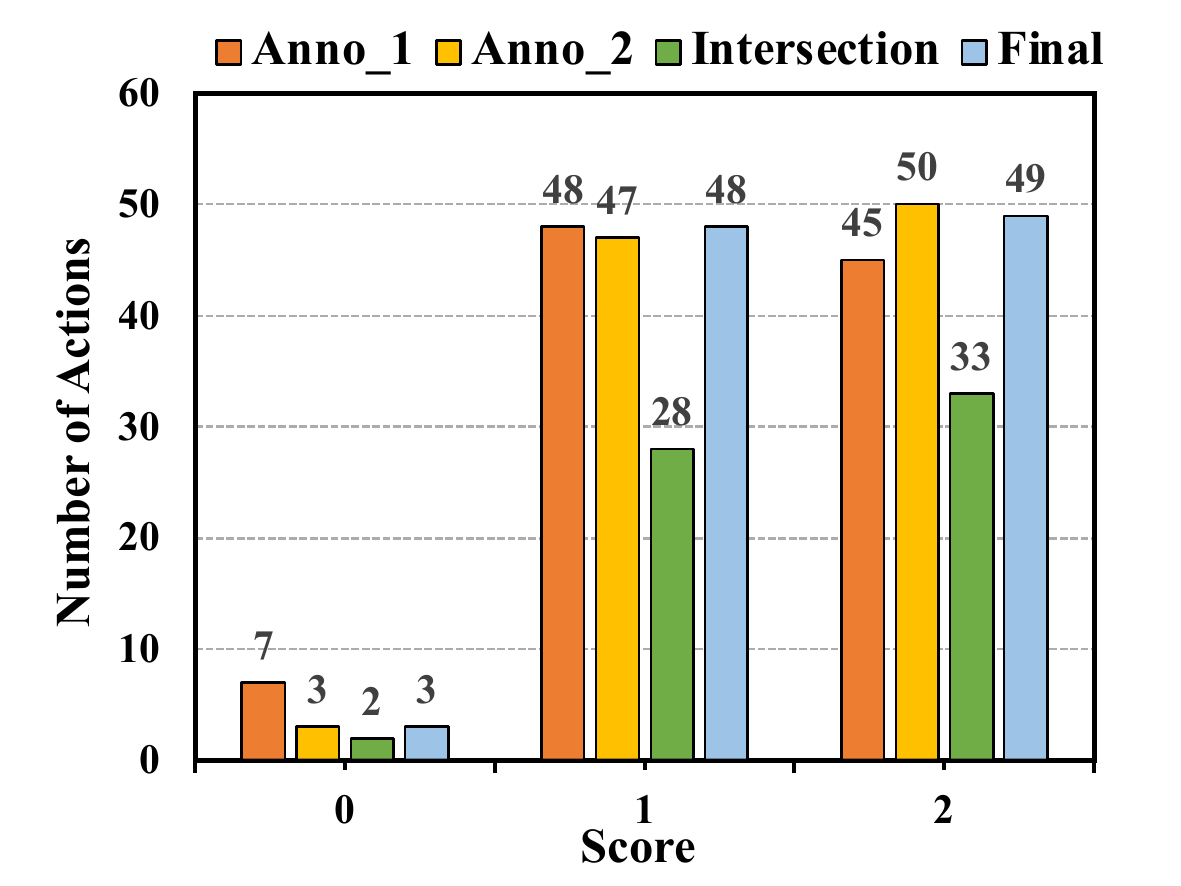}
\caption{Human evaluation results for automatically extracted error templates.}
\label{fig:dis1}
\end{figure}
In order to automatically select an appropriate corrective action for a specific template, we use the language model perplexity (PPL)  as a criterion. PPL is defined as the exponentiated average negative log-likelihood of a sequence. A sentence is more likely to be grammatical when it has a lower PPL. For a tokenized sentence $X=(x_1, x_2, ..., x_t)$, the PPL can be computed as:
\begin{equation}
    \mbox{PPL}(X)=\mbox{exp}\{-\frac{1}{t}\sum_{i}^{t}\mbox{log}p_\theta(x_i|x_{<i})\}
\end{equation}
where $x_{<i}$ denotes the preceding tokens of the i-th token $x_i$, and $\theta$ refers to the parameters of the language model. 

We use each template to match $N$ sentences from a large-scale corpus\footnote{We use the Weixin public corpus: \url{https://github.com/nonamestreet/weixin_public_corpus}.} and perform corrective actions on them. Then we measure the average PPL reduction ($\triangledown\mbox{PPL}$) to decide the golden action (ACT$_{g}$) for this template, as shown below:
\begin{equation}
    \triangledown\mbox{PPL}=\frac{1}{N}\sum_{i}^{N}(\mbox{PPL}_{b}(X_i) - \mbox{PPL}_{a}(X_i))
    \label{eq2}
\end{equation}

\begin{equation}
 \mbox{ACT}_{g}\!=\!\left\{
\begin{aligned}
 &\mbox{ACT}_{left},\,  \mbox{if} \, \triangledown\mbox{PPL}_{left}\!-\!\triangledown\mbox{PPL}_{right}\!>\!\alpha; \\
 &\mbox{ACT}_{right},\,  \mbox{elif} \, \triangledown\mbox{PPL}_{right}\!-\! \triangledown\mbox{PPL}_{left}\!>\!\alpha;  \\
 &\mbox{ACT}_{random},\,  \mbox{else.}
\end{aligned}
\right.
\label{eq3}
\end{equation}
where $\mbox{PPL}_{b}$ and $\mbox{PPL}_{a}$ denote the perplexity \textit{before} and \textit{after} performing a specific corrective action, respectively. $\triangledown\mbox{PPL}_{left}$ means the average PPL reduction after performing ACT$_{left}$ and the same goes to $\triangledown\mbox{PPL}_{right}$. $\alpha > 0$ is a hyperparameter.

\begin{figure}[tp!]
\centering
\includegraphics[scale=0.32]{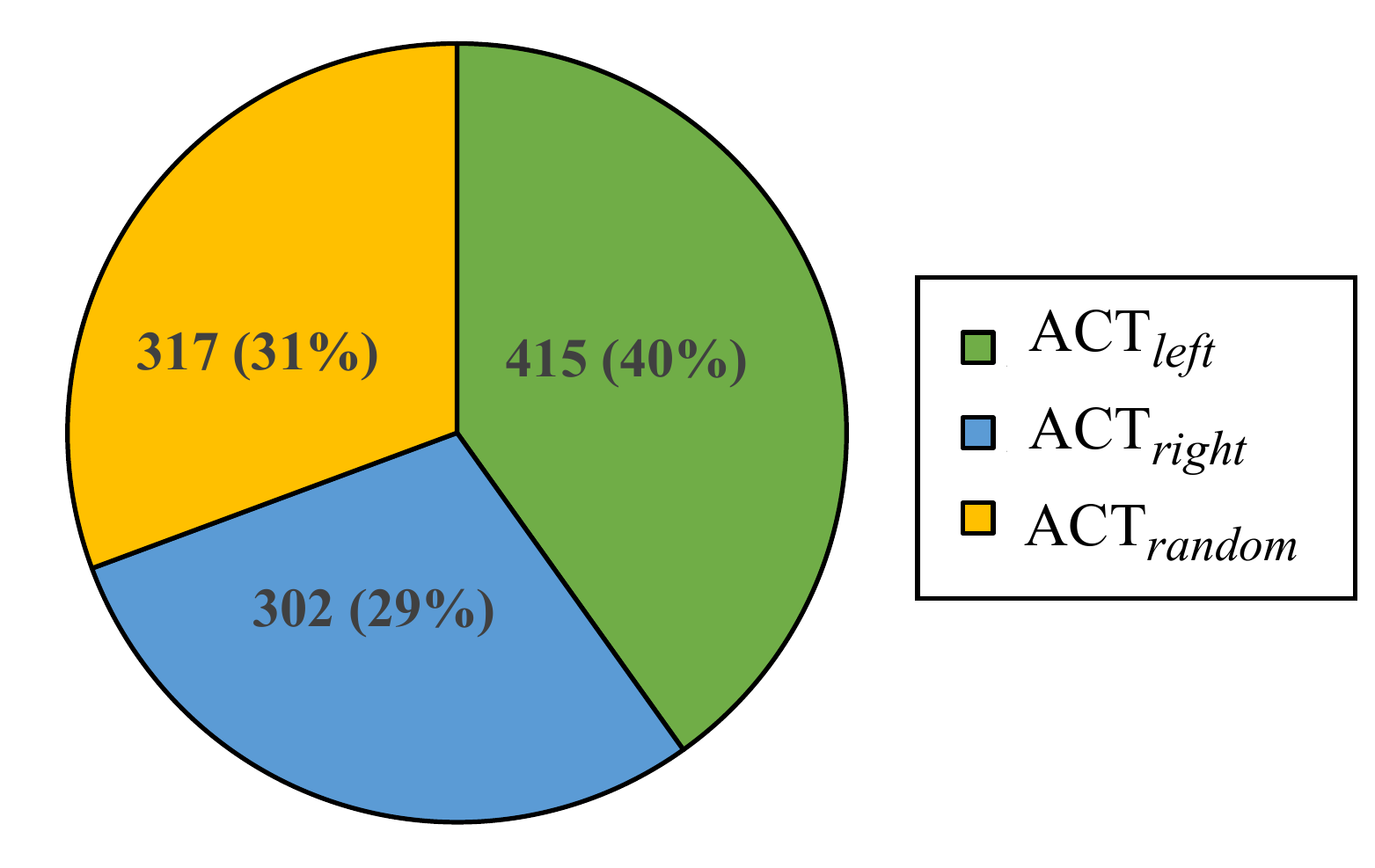}
\caption{The number and proportion of each kind of corrective action for our all 1,119 error templates.
}
\label{fig:p}
\end{figure}

In practical experiments, we set the hyperparameter $N$ in Equation \ref{eq2} to 20 and $\alpha$ in Equation \ref{eq3} to 5, and use Chinese-GPT-2 language model\footnote{\url{https://github.com/Morizeyao/GPT2-Chinese}} \citep{radford2019language} to compute the perplexity. Figure \ref{fig:p} presents the numbers and proportions of derived corrective actions for our 1,119 templates, among which ACT$_{left}$ is the most frequent.

\paragraph{Human Evaluation.} We further ask the annotators to assess the quality of the automatically derived corrective actions. We discard the 0-point error templates in Figure \ref{fig:dis1} and use the rest 92 templates for evaluation. The acceptability of actions follows a 3-point scale similar to the human evaluation for templates:
\vspace{-3mm}
\begin{itemize} 
    \setlength{\itemsep}{-2pt}
    \setlength{\topsep}{-2pt}
    \item \textbf{[0]}: the action can \textit{hardly} fix the error detected by the template correctly;
    \item \textbf{[1]}: the action can \textit{sometimes} fix the error detected by the template correctly;
    \item \textbf{[2]}: the action can \textit{always} fix the error detected by the template correctly;
\end{itemize}
\vspace{-3mm}

\begin{figure}[tp!]
\centering
\includegraphics[scale=0.55]{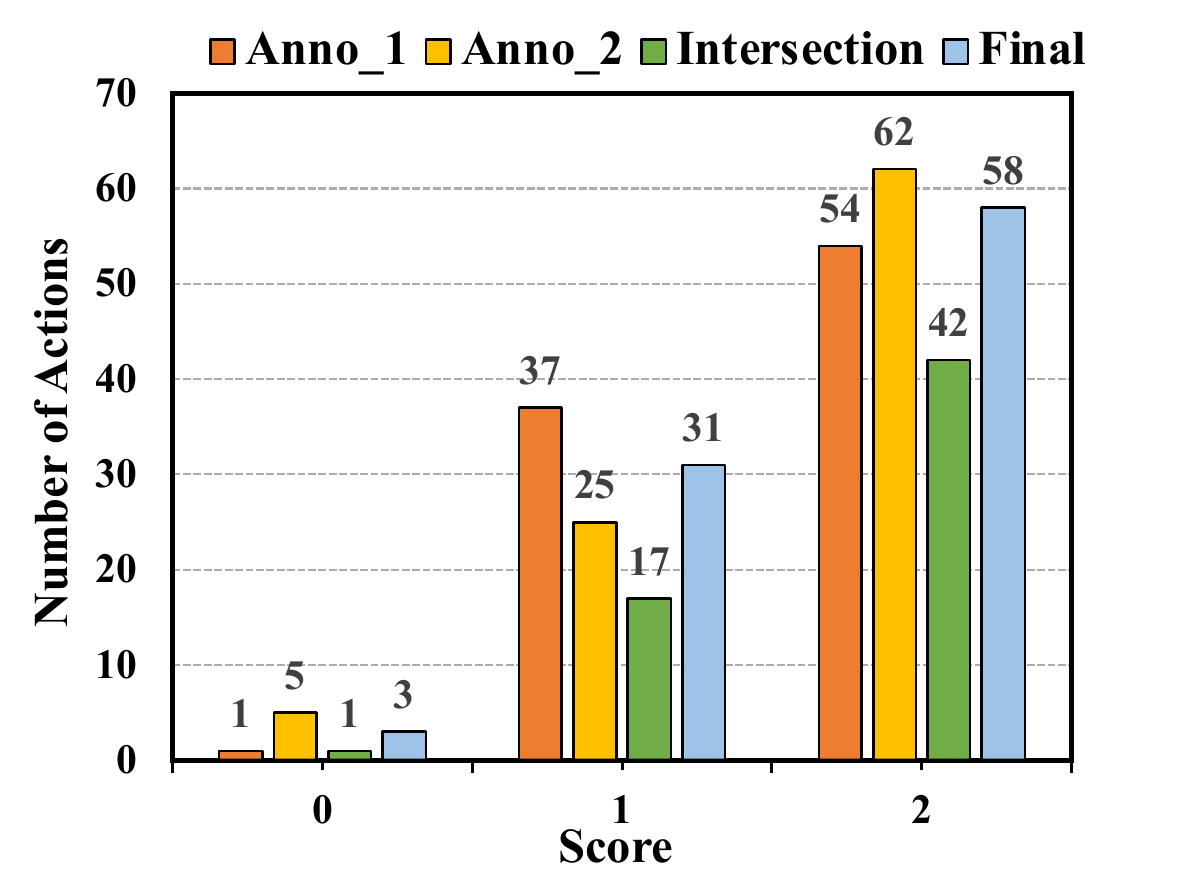}
\caption{Human evaluation results for automatically selected corrective actions.}
\label{fig:dis2}
\end{figure}

We present the evaluation results in Figure \ref{fig:dis2}. The average scores for actions are 1.58 and 1.62 from two annotators with an annotation constituency ratio of 65.22\%. Similarly, annotators were asked to discuss inconsistent cases and determine a unique score for each case. At least 96.74\% of actions are considered effective in certain cases (score > 0) and 63.04\% of actions are considered always effective (score = 2). Ultimately, we believe that we have demonstrated the solid quality of our automatically acquired templates and actions.

\section{Experiment and Analysis}
In this section, we conduct experiments on several public Chinese GEC datasets to show the effectiveness of our proposed method. 

\subsection{Experimental Setup}

\paragraph{Datasets.}
We report results on the following Chinese GEC datasets: CTC-2021\footnote{Since the test set of CTC-2021 requires the online submission, we report results on the qualification set for simplicity.}, NLPCC-2018 \citep{zhao2018overview}\footnote{\url{http://tcci.ccf.org.cn/conference/2018/taskdata.php}}, and MuCGEC \citep{zhang2022mucgec}\footnote{\url{https://github.com/HillZhang1999/MuCGEC}}. NLPCC-2018 and MuCGEC focus on texts written by Chinese-as-a-Second-Language learners. CTC-2021 instead considers texts written by native speakers and further includes some errors that rarely occur in learner texts. 

\paragraph{Baselines.}
In terms of baselines, we directly use the state-of-the-art GEC system described in \citet{zhang2021ctcreport} for CTC-2021. For the other two benchmarks, we follow the training setting from \citet{zhang2022mucgec} and build a strong sequence-to-edit GEC model. 
To make use of error templates, we leverage them to match potentially erroneous spans in all input sentences and perform the golden corrective actions before feeding these sentences into downstream GEC models.

\subsection{Main Results}
We present the main experimental results on Chinese GEC benchmarks in Table \ref{tab:res}. On the CTC-2021 benchmark, incorporating our error templates leads to a substantial improvement of 4.50 correction-level F$_1$ value compared with the baseline system, which confirms the effectiveness of our proposed method. When only using error templates, the P/R/F value on CTC-2021 is 64.29/12.12/20.39, which demonstrates that our templates already cover a considerable number of errors.

Specifically, using error templates improve the recall value of the baseline from 9.59 and 7.41 to 63.01 and 40.74 for two error types defined by the benchmark organizers: \textit{meaning duplicates} and \textit{mixed sentence patterns} (see Table \ref{tab:example}). Both types fit our templates well and can be fixed by performing delete operations. Moreover, both types are frequently asked in the Q\&A platforms, so our templates have relatively high coverage of them.

However, when experimenting on NLPCC-18 and MuCGEC, the improvement from using our templates seems to be non-existent. Such phenomena can be explained as NLPCC-18 and MuCGEC are from learner texts, while our templates are collected from the native text source. In fact, only 0.55\% and 0.43\% of sentences are modified after using error templates on NLPCC-18 and MuCGEC, while the proportion is 10.1\% on CTC-2021.

\definecolor{Gray}{gray}{0.9}
\begin{table}[tp!]
\centering
\scalebox{1.0}{
\begin{tabular}{cccc}
\hline
                       & \textbf{P} & \textbf{R} & \textbf{F} \\ \hline
\multicolumn{4}{c}{\textbf{CTC-2021}}                         \\ \hline
\textbf{Only ET}      & \textbf{64.29}      & 12.12      & 20.39      \\
\textbf{Baseline}      & 51.12      & 58.79      & 54.71      \\ \rowcolor{Gray}
\textbf{Baseline + ET} &52.87      & \textbf{67.28}      & \textbf{59.21}      \\ \hline\multicolumn{4}{c}{\textbf{NLPCC-2018}}                         \\ \hline
\textbf{Baseline}      & \textbf{43.12}      & 30.18      & \textbf{39.72}      \\\rowcolor{Gray}
\textbf{Baseline + ET} & 43.07      & \textbf{30.22}      & 39.69     \\ \hline\multicolumn{4}{c}{\textbf{MuCGEC}}                         \\ \hline
\textbf{Baseline}      & \textbf{44.65}      & 27.32      & \textbf{39.62}      \\\rowcolor{Gray}
\textbf{Baseline + ET} & 44.56      &\textbf{ 27.37}      & 39.59      \\ \hline
\end{tabular}
}
\caption{Results on three Chinese GEC benchmarks. ``ET'' refers to Error Template. ``P'', ``R'', ``F'' denote Precision, Recall and F-value, respectively. We use the F$_{1}$ metric for CTC-2021 and F$_{0.5}$ metric for the other two benchmarks according to their official settings.}
\label{tab:res}
\end{table}

\subsection{Analysis}

\paragraph{Effectiveness of Perplexity-based Corrective Action Selection.} To figure out whether it is necessary to utilize the perplexity to select the corrective actions of error templates, we conduct an ablation study on CTC-2021. Concretely, we try to delete the left parts of all templates (All left), delete the right parts of all templates (All right) and randomly delete one side of all templates (All random) separately. As shown in Table \ref{tab:ab}, using the corrective actions selected based on perplexity (Baseline + ET) significantly outperforms other methods, which clearly demonstrates its effectiveness and necessity.


\paragraph{Impact of the Timing of Using Templates.} As discussed before, we use error templates to correct sentences in the pre-processing stage before feeding them into GEC models. However, we can also use templates in the post-processing stage after performing corrections using GEC models. Table \ref{tab:time} present the results of this comparative experiment on CTC-2021. ``Both'' means that using templates in both stages. We can see that using templates in the pre-processing stage achieves the best performance.

\begin{table}[tp!]
\centering
\scalebox{1.0}{
\begin{tabular}{lccc}
\hline
\multicolumn{1}{c}{}   & \textbf{P} & \textbf{R} & \textbf{F$_1$} \\ \hline
\textbf{Baseline + ET} & \textbf{52.87}      & \textbf{67.28 }     & \textbf{59.21}       \\
\textbf{\hspace{0.3cm} - All Left}    & 50.15      & 63.46      & 56.03       \\
\textbf{\hspace{0.3cm} - All Right}   & 48.48      & 61.15      & 54.08       \\
\textbf{\hspace{0.3cm} - All Random}  & 49.47      & 62.50      & 55.23       \\ \hline
\end{tabular}
}
\caption{Ablation study on the perplexity-based corrective action selection on CTC-2021.}
\label{tab:ab}
\end{table}
\begin{table}[tp!]
\centering
\scalebox{1.0}{
\begin{tabular}{lccc}
\hline
\multicolumn{1}{c}{}   & \textbf{P} & \textbf{R} & \textbf{F$_1$} \\ \hline
\textbf{Pre-process} & \textbf{52.87}      & \textbf{67.28 }     & \textbf{59.21}       \\
\textbf{Post-process} & 52.51      & 66.35     & 58.62       \\
\textbf{Both} &   52.74    & 67.14     & 59.08       \\
\hline
\end{tabular}
}
\caption{Effect of the timing of applying error templates.}
\label{tab:time}
\end{table}

\paragraph{Frequency of Use of Templates.} To investigate how frequently our templates are used, we count their frequency of use on CTC-2021. 86 templates are used to perform 98 corrections on CTC-2021. Among them, 76 (88.4\%) templates are only used once and 9 (10.4\%) templates are used twice. Only 1 (1.2\%) templates are used for three times. Such a phenomenon indicates that errors corrected by our templates are basically long-tail, which models often fail to repair since they are very sparse in training data. This observation supports that our templates can be a good complement to models.

\begin{table*}[tp!]
\centering
\scalebox{1.0}{
\begin{tabular}{lll}
\hline
\textbf{Baseline}      & \begin{tabular}[c]{@{}l@{}}要形成这么大一座环礁，\textcolor{blue}{至少}需要百万年\textcolor{blue}{以上}的时间。\\ It will take \textcolor{blue}{at least} \textcolor{blue}{more than} millions of years to build such a large atoll.\end{tabular} & \xmark \\
\textbf{Baseline + ET} & \begin{tabular}[c]{@{}l@{}}要形成这么大一座环礁，至少需要百万年的时间。\\ It will take at least millions of years to build such a large atoll.\end{tabular}           & \cmark \\ \hline
\textbf{Baseline}      & \begin{tabular}[c]{@{}l@{}}\textcolor{blue}{如果}说郭沫\textcolor{blue}{若是}一位伟大的文学大师。\\ \textcolor{blue}{If} we say that Guo Mo\textcolor{blue}{ruo is} a great literary master.\end{tabular}                                  & \cmark \\
\textbf{Baseline + ET} & \begin{tabular}[c]{@{}l@{}}如果说郭沫一位伟大的文学大师。\\ If we say that Guo Mo a great literary master.\end{tabular}                                          & \xmark \\ \hline
\end{tabular}
}
\caption{Comparison of results using error templates or not. The matched templates are highlighted in \textcolor{blue}{blue}.}
\label{tab:case}
\end{table*}

\paragraph{Case Study.} We list two cases before and after incorporating our error templates in Table \ref{tab:case}. In the first case, we can see that the error template locates and corrects the error successfully. However, in the second case, the error template modifies a correct sentence improperly. This observation is quite interesting and can disclose a flaw of our method --- the error templates represented as regular expressions inevitably ignore the contextual information. For example, ``若是'' in the template ``如果.*若是'' should mean ``if'', while in the second case, ``若'' is a character in a Chinese name and ``是'' means ``is'', which leads to the improper modification. We will attempt to make some improvements in this direction in our future work.

\section{Conclusion}
This paper presents a rule-based GEC method based on automatically obtained error templates. We first use web crawlers to extract candidate error templates conforming to pre-defined search patterns, then employ the language model perplexity as a criterion to select a proper corrective action for each error template. The experimental results on the CTC-2021 Chinese GEC dataset confirm the effectiveness of our method. In the future, we plan to extend our method to other languages and cover more error templates that need other actions like substitution and insertion to correct.


\bibliography{anthology,custom}
\bibliographystyle{acl_natbib}




\end{CJK}
\end{document}